\documentclass[10pt,twocolumn,letterpaper]{article}

\usepackage{cvpr}
\usepackage{times}
\usepackage{graphicx}
\usepackage{amsmath}
\usepackage{amssymb}
\usepackage{mathtools, amssymb, amsthm, bbm}
\usepackage{enumitem}

\usepackage[numbers]{natbib}
\usepackage{color}
\usepackage{verbatim}

\usepackage{tikz}
\usetikzlibrary{shapes,snakes,positioning,arrows,calc,fit}

\newcommand{\Sc}{\mathcal{S}}
\newcommand{\X}{\mathcal{X}}
\newcommand{\Y}{\mathcal{Y}}
\newcommand{\br}[1]{\left({#1}\right)}
\newcommand{\eos}{\mathbf{eos}}

\newcommand{\ignore}[1]{}
\usepackage{amsmath}
\DeclareMathOperator*{\argmax}{argmax}

\usepackage[utf8x]{inputenc} 

    \makeatletter
\def\@fnsymbol#1{\ensuremath{\ifcase#1 \or 	\spadesuit\or \clubsuit \or \dagger\or \ddagger\or
   \mathsection\or \mathparagraph\or \|\or **\or \dagger\dagger
   \or \ddagger\ddagger \else\@ctrerr\fi}}
    \makeatother

\usepackage[breaklinks=true,bookmarks=false]{hyperref}

\cvprfinalcopy % *** Uncomment this line for the final submission

\def\cvprPaperID{5817} % *** Enter the CVPR Paper ID here

\begin{document}

%%%%%%%%% TITLE
\title{Knowing When to Stop: \\ Evaluation and Verification of Conformity to Output-size Specifications}

\author{Chenglong Wang\thanks{work done during an internship at DeepMind}\\
University of Washington\\
{\tt\small clwang@cs.washington.edu}
% For a paper whose authors are all at the same institution,
% omit the following lines up until the closing ``}''.
% Additional authors and addresses can be added with ``\and'',
% just like the second author.
% To save space, use either the email address or home page, not both
\and
Rudy Bunel\footnotemark[1]\\
University of Oxford\\
{\tt\small rudy@robots.ox.ac.uk}
\and
Krishnamurthy Dvijotham\\
DeepMind\\
{\tt\small dvij@google.com}
\and
Po-Sen Huang\\
DeepMind\\
{\tt\small posenhuang@google.com}
\and
Edward Grefenstette\thanks{now at Facebook AI Research}\\
DeepMind\\
{\tt\small egrefen@fb.com}
\and
Pushmeet Kohli\\
DeepMind\\
{\tt\small pushmeet@google.com}
}

\maketitle

\begin{abstract}
Models such as {\em Sequence-to-Sequence} and {\em Image-to-Sequence} are widely used in real world applications. While the ability of these neural architectures to produce variable-length outputs makes them extremely effective for problems like Machine Translation and Image Captioning, it also leaves them vulnerable to failures of the form where the model produces outputs of undesirable length. This behaviour can have severe consequences such as usage of increased computation and induce faults in downstream modules that expect outputs of a certain length. Motivated by the need to have a better understanding of the failures of these models, this paper proposes and studies the novel output-size modulation problem and makes two key technical contributions. First, to evaluate model robustness, we develop an easy-to-compute differentiable proxy objective that can be used with gradient-based algorithms to find output-lengthening inputs. Second and more importantly, we develop a verification approach that can formally verify whether a network always produces outputs within a certain length. Experimental results on Machine Translation and Image Captioning show that our output-lengthening approach can produce outputs that are $50$ times longer than the input, while our verification approach can, given a model and input domain, prove that the output length is below a certain size. 
\end{abstract}

\section{Introduction}

Neural networks with variable output lengths have become ubiquitous in several applications. In particular, recurrent neural networks (RNNs) such as LSTMs \citep{hochreiter1997long}, used to form ``sequence'' models \citep{sutskever2014sequence}, have been successfully and extensively applied in in image captioning \citep{vinyals2015show, DBLP:conf/acl/SennrichHB16a,DBLP:journals/pami/KarpathyF17,ChenZ15,FangGISDDGHMPZZ15,XuBKCCSZB15,imc_rennie2017self,imc_lu2017knowing,imc_Anderson_2018_CVPR}, video captioning~\citep{vdc_venugopalan2015sequence,vdc_xu2015jointly,vdc_wang2018bidirectional,vdc_yao2015describing,vdc_yu2016video}, machine translation (MT) \citep{sutskever2014sequence, cho2014learning}, summarization \citep{chopra2016abstractive}, and in other sequence-based transduction tasks.

%However, due to the black-box nature of these approaches, they are vulnerable to undesirable behavior \citep{jia2017adversarial, ribeiro2018semantically}.  While these previous works demonstrate that the input can be modified to produce undesirable outputs, the resulting outputs can be ad-hoc and there is no clear understanding of the property that causes outputs to be undesirable, which makes it difficult to design systematic attacks or perform  verification to prove the absence of these attacks.

% \djred{Need to add more here \citep{papernot2016crafting}, show and tell, seq2sick, SEAR, hotflip}. 

% \djred{conform output size specs
% specs - output length/input length ratio
% not conform - adversarial attack
% conform - be able to verify the specs. 
% }
%While these examples expose undesirable behavior in variable compute models, they are not fundamentally distinct from adversarial examples in image classification since they study perturbations of the input that cause incorrect predictions. However, models with variable-length output are susceptible to another kind of vulnerability. It is possible that there are inputs that cause the models to produce unnecessarily long outputs that are irrelevant to the task being solved. 

The ability of these sequence neural models to generate variable-length outputs is key to their performance on complex prediction tasks. However,
this ability also opens a powerful attack for adversaries that try to force the model to produce outputs of specific lengths that, for instance, lead to increased computation or affect the correct operation of down-stream modules. To address this issue, we introduce the \emph{output-length modulation} problem where given a {\em specification} of the form that the model should produce outputs with less than a certain maximum length, we want to find adversarial examples, {\em i.e.} search for inputs that lead the model to produce outputs with a larger length and thus show that the model under consideration violates the specification. Different from existing work on targeted or untargeted attacks where the goal is to perturb the input such that the output is another class or sequence in the development dataset (thus within the dataset distribution), the output-modulation problem requires solving a more challenging task of finding inputs such that the output sequences are outside of the training distribution, which was previously claimed difficult~\cite{show_and_fool}.

The naive approach to the solution of the output-length modulation problem involves a computationally intractable search over a large discrete search space. To overcome this, we develop an easy-to-compute differentiable proxy objective that can be used with gradient-based algorithms to find output-lengthening inputs. Experimental results on Machine Translation show that our adversarial output-lengthening approach can produce outputs that are $50$ times longer than the input. However, when evaluated on the image-to-text image captioning model, the method is less successful. There could have been two potential reasons for this result: 
the image-to-text architecture is truly robust, or the adversarial approach is not powerful enough to find adversarial examples for this model.  
To resolve this question, we develop a verification method for checking and formally proving whether a network is consistent with the output-size specification for the given range of inputs. 
To the best of our knowledge, our verification algorithm is the first formal verification approach to check properties of recurrent models with variable output lengths.

\paragraph{Our Contributions} To summarize, the key contributions of this paper are as follows:
\begin{itemize}\itemsep-3pt
    \item We propose and formulate the novel output-size modulation problem to study the behaviour of neural architectures capable of producing variable length outputs, and we study its evaluation and verification problems. 
    \item For evaluation, we design an efficiently computable differentiable proxy for the expected length of the output sequence. Experiments show that this proxy can be optimized using gradient descent to efficiently find inputs causing the model to produce long outputs. 
    \item We demonstrate that popular machine translation models can be forced to produce long outputs that are 50 times longer than the input sequence. The long output sequences help expose modes that the model can get stuck in, such as undesirable loops where they continue to emit a specific token for several steps.
    \item We demonstrate the feasibility of formal verification of recurrent models by proposing the use of mixed-integer programming to formally verify that a certain neural image-captioning model will be consistent with the specification for the given range of inputs.
\end{itemize}

\paragraph{Motivations and Implications}
Our focus on studying the output-length modulation problem is motivated by the following key considerations:
\begin{itemize}\itemsep-1pt
    \item \emph{Achieving Computational Robustness:} Many ML models are now offered as a service to customers via the cloud. In this context, ML services employing variable-output models could be vulnerable to denial-of-service attacks that cause the ML model to perform wasteful computations by feeding it inputs that induce long outputs. This is particularly relevant for \emph{variable compute models}, like Seq2Seq  \citep{cho2014learning, sutskever2014sequence}. Given a trained instance of the model, no method is known to check for the consistency of the model with a specification on the number of computation steps. Understanding the vulnerabilities of ML models to such output-lengthening and computation-increasing attacks is important for the safe deployment of ML services.  
    
    \item \emph{Understanding and Debugging Models:} By designing inputs that cause models to produce long outputs, it is possible to reason about the internal representations learned by the model and isolate where the model exhibits undesirable behavior. For example, we find that an English to German sequence-to-sequence model can produce outputs that end with a long string of question marks (`?'). This indicates that when the output decoder state is conditioned on a sequence of `?'s, it can end up stuck in the same state.
    
    \item \emph{Uncovering security vulnerabilities through adversarial stress-testing:} 
    The adversarial approach to output-length modulation tries to find parts of the space of inputs where the model exhibits improper behavior. Such inputs does not only reveal abnormal output size, but could also uncover other abnormalities like the privacy violations of the kind that were recently revealed by \citep{carlini_secret} where an LSTM was forced to output memorized data.
    
    \item\emph{Canonical specification for testing generalization of variable-output models:} Norm-bounded perturbations of images \citep{szegedy2013intriguing} have become the standard specification to test attacks and defenses on image classifiers. While the practical relevance of this particular specification can be questioned \citep{goodfellow_question}, it is still served as a useful canonical model encapsulating the essential difficulty in developing robust image classifiers. We believe stability of output-lengths can serve a similar purpose: as a canonical specification for variable output-length models. 
    The main difficulties in studying variable output length models in an adversarial sense (the non-differentiability of the objective with respect to inputs) are exposed in output-lengthening attack, making it a fertile testing ground for both evaluating attack methods and defenses. We hope that advances made here will facilitate the study of robustness on \emph{variable compute models} and other specifications for variable-output models such as monotonicity.
\end{itemize}

%Just like norm bounded perturbations have been a useful canonical model for adversarial examples in classifiers, we hope that output-lengthening can be viewed as another class of canonical attacks that could be studied across a variety of models with variable output size.

\section{Related Work} 
% Adverserial attack
% \paragraph{Adversarial attack}
%In image classification tasks, researchers have observed that even state-of-the-art models can be fooled in changing their predictions by making small but carefully chosen modifications to the input, known as adversarial perturbations
%\citep{szegedy2013intriguing, kurakin2016adversarial_2, carlini2017adversarial, carlini2017towards}.

% Related to seq2sick, show and fool, SEARs, hotflip
There are several recent studies on generating adversarial perturbations on variable-output models. \cite{ribeiro2018semantically,jia2017adversarial} show that question answering and machine comprehension models are sensitive to attacks based on semantics preserving modification or the introduction of unrelated information. 
\cite{ebrahimi2017hotflip,yang2018greedy} find that character-level classifiers are highly sensitive to small character manipulations. 
\cite{ShekharPKHNSB17} shows that models predicting the correctness of image captions struggle against perturbations consisting of a single word change. 
\cite{show_and_fool} and \cite{seq2sick} further study adversarial attacks for sequential-output models (machine-translation, image captioning) with specific target captions or keywords. 
% Some recent work in shows that other NLP models are also sensitive to input perturbations (seq2sick, show and fool, SEARs, hotflip). 
% We study whether a model can conform with the output size specification.

We focus on sequence output models and analyze the output-length modulation problem, where the models should produce outputs with at least a certain number of output tokens. 
We study whether a model can be adversarially perturbed to change the size of the output, which is a more challenging task compared to targeted attacks (see details in Section \ref{sec:formulation}). On the one hand, existing targeted attack tasks aim to perturb the input such that the output is another sequence in the validation dataset (thus within the training distribution), but attacking output size requires the model to generate out-of-distribution long sequences. On the other hand, since the desired output sequence is only loosely constrained by the length rather than directly provided by the user, the attack algorithm is required to explore the output size to make the attack possible. 

For models that cannot be adversarially perturbed, we develop a verification approach to show that it isn't simply a lack of power by the adversary but the sign of true robustness from the model. Similar approaches have been investigated for feedforward networks \citep{bunel2018piecewise, cheng2017maximum, tjeng2017verifying} but our work is the first to handle variable output length models and the corresponding decoding mechanisms.

% and for fixed input recurrent neural networks \citep{rnn_verif} 

% Possibly mention " Verification of Continuous Time Recurrent Neural Networks (Benchmark Proposal)", but they don't provide solution, just a problem, and this is for continuous time recurrent neural networks, which is not something practically used.

\section{Modulating Output-size}
\label{sec:formulation}

We study neural network models capable of producing outputs of variable length. We start with a canonical abstraction of such models, and later specialize to concrete models used in machine translation and image captioning.

We denote by $x$ the input to the network and by $\X$ the space of all inputs to the network. We consider a set of inputs of interest $\Sc$, which can denote, for example, the set of ``small''\footnote{The precise definition of small is specific to the application studied.} perturbations of a nominal input. We study models that produce variable-length outputs sequentially. Let $y_t\in \Y$ denote the $t$-th output of the model, where $\Y$ is the output vocabulary of the model. 
At each timestep, the model defines a probability over the next element $P(y_{t+1} | x, y_{0:t})$. There exists a special end-of-sequence element $\eos \in \Y$ that signals termination of the output sequence.

In practice, different models adopt different decoding strategies for generating $y_{t+1}$ from the probability $P(y_{t+1}|x,y_{0:t})$ 
\citep{germann2001fast,jelinek1997statistical, koehn2003statistical}. In this paper, we focus on the commonly used deterministic \emph{greedy decoding} strategy \citep{germann2001fast}, where at each time step, the generated token is given by the argmax over the logits:
\begin{subequations}
\begin{align}
y_0 &= \argmax\left\{P\br{\cdot | x}\right\} \\    
y_{t+1} &= \argmax \left\{ P\br{\cdot | x, y_{0:t}}\right\} \text{ if } y_t \neq \eos
\end{align}
\end{subequations}
Since greedy decoding is deterministic, for a given sample $x$ with a finite length output, we can define the length of the greedily decoded sequence as:
\begin{equation}
    \ell\br{x}\!=\!t\ \text{ s.t } \begin{array}{l} y_t = \eos\\ y_i \neq \eos\quad\forall i < t \\
    y_{i+1} = \argmax\left\{P\left(.| x, y_{0:i}\right)\right\} \quad \forall i < t\end{array}\hspace{-.5em}
    \label{eq:greedy-length}
\end{equation}
Note that there is a unique $t$ that satisfies the above conditions, which is precisely the first $t$ at which $y_t=\eos$ when using greedy decoding.

\paragraph{Output length modulation specification}
A network is said to satisfy the \emph{output length modulation specification} parameterized by $\Sc, \hat{K}$, if for all inputs $x$ in $\Sc$, the model terminates within $\hat{K}$ steps under greedy decoding for all $x \in \Sc$, formally:
\begin{align}
\forall x \in \Sc\quad \quad \ell(x) \leq \hat{K}
\label{eq:spec_output_gd}
\end{align}\vspace{-20pt}
In Section \ref{sec:perturbations}, we study the problem of finding adversarial examples, i.e., searching for inputs that lead the model to produce outputs with a larger length and thus show that the model violates the specification.
In Section \ref{sec:verifications}, we use formal verification method to prove that a model is consistent with the specification for the given range of inputs, if such attacks are indeed impossible.

\subsection{The Output-Size Modulation Problem}
\label{sec:perturbations}

In order to check whether the specification, Eq.~\eqref{eq:spec_output_gd}, is valid, one can consider a falsification approach that tries to find counterexamples proving that Eq.~\eqref{eq:spec_output_gd} is false. If an exhaustive search over $\Sc$ for such counterexamples fails, the specification is indeed true. However, exhaustive search is computationally intractable; hence, in this section we develop gradient based algorithms that can efficiently find counterexamples (although they may miss them even if they exist). To develop the falsification approach, we study the solution to the following optimization objective:
\begin{align}
    \max_{x \in \Sc}\ \ell\br{x}
\label{eq:atk_objective}
\end{align}
where $\Sc$ is the valid perturbation space. If the optimal solution $x$ in the space $\Sc$ has $\ell(x) >\hat{K}$, then \eqref{eq:spec_output_gd} is false. 

The attack spaces $\Sc$ we consider in this paper include both {continuous inputs} (for image-to-text models) and {discrete inputs} (for Seq2Seq models).

\smallskip

\emph{Continuous inputs}:
For continuous inputs, such as image captioning tasks, the input is an $n \times m$ image with pixel values normalized to be in the range $[-1, 1]$. $x$ is an $n \times m$ matrix of real numbers and $\X=[-1,1]^{n\times m}$. We define the perturbation space $\Sc(x,\delta)$ as follows:
\[\Sc(x,\delta)=\{x'\in\X \mid \|x'-x\|_\infty \le \delta\}\]
i.e., the space of $\delta$ perturbations of the input $x$ in the $\ell_\infty$ ball.

\smallskip

\emph{Discrete inputs}:
For discrete inputs, e.g., machine translation tasks, inputs are discrete tokens in a language vocabulary. Formally, given the vocabulary $V$ of the input language, the input space $\X$ is defined as all sentences composed of tokens in $V$, i.e., $\X=\{(x_1,\dots, x_n)\mid x_i\in V, n>0\}$. 
Given an input sequence $x=(x_1,\dots,x_n)$, we define the $\delta$-perturbation space of a sequence as all sequences of length $n$ with at most $\lceil\delta\cdot n\rceil$ tokens different from $x$ (i.e., $\delta\in[0, 1]$ denotes the percentage of tokens that an attacker is allowed to modify). Formally, the perturbation space $\Sc(x,\delta)$ is defined as follows:

\mbox{\small $\Sc(x,\delta)=\{(x'_1,\dots,x'_n) \in V^n \mid \sum\limits_{i=1}^{n}\mathbbm{1}[x_i \neq x'_i] \le \lceil\delta\cdot n\rceil\}$}

%Since the input vocabulary $V$ is finite, the perturbation space is a finite discrete space with size $|\Sc(x,\delta)|= {n\choose \delta\cdot n}|V|^{\delta\cdot n}$. %Note since the vocabulary size is often very large in practice, exhaustively search for the optimum attack is often computationally intractable.

\subsection{Extending Projected Gradient Descent Attacks}
\label{sec:pgd}
% Directly optimizing Eq. \eqref{eq:atk_objective} is not tractable due to the fact that (1) it is an infinite sum with each term being itself the sum over a combinatorially large number of paths; and (2) the input space of the discrete input sequences is exponentially large. 
%  and then show how we extend it for output-lengthening attacks.

In the projected gradient descent (PGD) attacks \citep{madry2017towards},\footnote{Here the adversarial objective is stated as maximization, so the algorithm is Projected Gradient \emph{Ascent}, but we stick with the PGD terminology since it is standard in the literature} given an objective function $J(x)$, the attacker calculates the adversarial example by searching for inputs in the attack space to maximize $J(x)$. In the basic attack algorithm, we perform the following updates at each iteration:
\begin{align}
    x' = \Pi_{{\Sc}(x,\delta)}\left(x + \alpha \nabla_x J(x)\right) \label{eq:pgd}
\end{align}

\noindent where \mbox{$\alpha >0$} is the step size and $\Pi_{{\Sc}(x,\delta)}$ denotes the projection of the attack to the valid space $\Sc(x,\delta)$. 
Observe that the adversarial objective in Eq.~\eqref{eq:atk_objective} cannot be directly used as $J(x)$ to update $x$ as the length of the sequence is not a differentiable objective function. This hinders the direct application of PGD to output-lengthening attacks.  Furthermore, when the input space $\Sc$ is discrete, gradient descent cannot be directly be used because it is only applicable to continuous input spaces.

In the following, we show our extensions of the PGD attack algorithm to handle these challenges.

\paragraph{Greedy approach for sequence lengthening}
\label{sec:greedy_relaxation}
We introduce a differentiable proxy of $\ell(x)$. Given an input $x$ whose decoder output logits are $(o_1,\dots,o_k)$ (i.e., the decoded sequence is $y=(\argmax(o_1),\dots,\argmax(o_k))$), instead of directly maximizing the output sequence length, we use a greedy algorithm to find an output sequence whose length is longer than $k$ by minimizing the probability of the model to terminate within $k$ steps. In other words, we minimize the log probability of the model to produce $\eos$ at any of the timesteps between $1$ to $k$. Formally, the proxy objective $\tilde{J}$ is defined as follows:
{
\[
\begin{array}{l}
\tilde{J}(x) = \sum\limits_{t=1}^k \max\left\{o_t[\eos] - \max\limits_{z\neq\eos}o_t[z],\ -\epsilon\right\}
\end{array}\]
}
where $\epsilon >0$ is a hyperparameter to clip the loss. This is piecewise differentiable w.r.t. the inputs $x$ (in the same sense that the ReLU function is differentiable) and can be efficiently optimized using PGD.
%Since evaluating the new proxy objective requires only running the encoder-decoder once, it can be efficiently calculated at each attack iteration in the greedy attack loops.

\subsection{Continuous relaxation for discrete inputs}
\label{sec:relaxation}

While we can apply the PGD attack with the proxy objective on the model with continuous inputs by setting the projection function $\Pi_{\Sc(x,\delta)}$ as the Euclidean projection, we cannot directly update discrete inputs. To enable a PGD-type  attack in the discrete input space, we use the Gumbel trick \citep{jang2016categorical} to reparameterize the input space to perform continuous relaxation of the inputs.

Given an input sequence $x=(x_1,\dots, x_n)$, for each $x_i$, we construct a distribution $\pi_i\in \mathbb{R}^{|V|}$ initialized with $\pi_i[x_i]=1$ and $\pi_i[z]=-1$ for all $z\in V \setminus\{x_i\}$. The softmax function applied to $\pi_i$ is a probability distribution over input tokens at position $i$ with a mode at $x_i$. With this reparameterization, instead of feeding $x=(x_1, \dots, x_n)$ into the model, we feed the Gumbel softmax sampling from the distribution $(u_1,\dots,u_n)$. The sample $\tilde{x}=(\tilde{x}_1,\dots,\tilde{x}_n)$ is calculated as follows:
\[
\begin{array}{cc}
 u_i \sim \text{Uniform}(0,1);\quad g_i = - \log( -\log(u_i)) \\
 p = \text{softmax}(\pi);\quad \tilde{x}_i=\text{softmax}(\frac{g_i + \log p_i}{\tau})
\end{array}
\]
where $\tau$ is the Gumbel-softmax sampling temperature that controls the discreteness of $\tilde{x}$. With this relaxation, we perform PGD attack on the distribution $\pi$ at each iteration. Since $\pi_i$ is unconstrained, the projection step in \eqref{eq:pgd} is unnecessary.

When the final $\pi'=(\pi'_1,\dots,\pi_n)$ is obtained from the PGD attack, we  draw samples $x_i'\sim \text{Categorical}(\pi_i)$ to get the final adversarial example for the attack.

\section{Verified Bound on Output Length}
\label{sec:verifications}
% We use formal verification if no attack is possible. %, which proves that the model will be consistent with the output-length specification for the given range of inputs, if no attack is possible.
% For the models that conform with the output-length specifications, we focus on using formal verification to conform the specification.
%TODO add introduction on why transition to verified bound
While heuristics approaches can be useful in finding attacks, they can fail due to the difficulty of optimizing nondifferentiable nonconvex functions. These challenges show up particularly when the perturbation space is small or when the target model is trained with strong bias in the training data towards short output sequences (e.g., the Show-and-Tell model as we will show in Section~\ref{sec:experiment}). Thus, we design a formal verification approach for complete reasoning of the output-size modulation problem, i.e., finding provable guarantees that no input within a certain set of interest can result in an output sequence of length above a certain threshold. 
% Our approach rely on counterexample search using complete methods. We encode optimization problem \eqref{eq:atk_objective} as a mixed-integer program (MIP) that is guaranteed to find the global optimum, albeit with a potential computational cost. %In this sense, they represent the strongest possible attacker in term of capabilities.

Our approach relies on counterexample search using intelligent brute-force search methods, taking advantage of powerful modern integer programming solvers \citep{GleixnerScip}. We encode all the constraints that an adversarial example should satisfy as linear constraints, possibly introducing additional binary variables. Once in the right formalism, these can be fed into an off-the-shelf Mixed Integer Programming (MIP) solver, which provably solves the problem, albeit with a potentially large computational cost. %In this sense, it represents the strongest possible attacker in term of capability.
The constraints consist of four parts: (1) the initial restrictions on the model inputs (encoding $\mathcal{S}(x, \delta)$), (2) the relations between the different activations of the network (implementing each layer), (3) the decoding strategy (connection between the output logits and the inputs at the next step), and (4) the condition for it being a counterexample ({\em ie}.~a sequence of length larger than the threshold). In the following, we show how each part of the constraints is encoded into MIP formulas.

Our formulation is inspired by pior work on encoding feed-forward neural networks as MIPs~\cite{bunel2018piecewise, cheng2017maximum, tjeng2017verifying}. The image captioning model we use consists of an image embedding model, a feedforward convolutional neural network that computes an embedding of the image, followed by a recurrent network that generates tokens sequentially starting with the initial hidden state set to the image embedding.

The image embedding model is simply a sequence of linear or convolutional layers and ReLU activation functions. Linear and convolutional layers are trivially encoded as linear equality constraints between their inputs and outputs, while ReLUs are represented by introducing a binary variable and employing the big-M method~\citep{grossmann2002review}:
\begin{subequations}
  \begin{flalign}
    x_i = \max\left(\hat{x}_i, 0\right) \ \Rightarrow \ &
    \delta_i \in \{0,1\},
    \quad x_i \geq 0\\
    & x_i \leq u_i \cdot \delta_i,
    \quad x_i\geq \hat{x}_i\\
    & x_i \leq \hat{x}_i - l_i\cdot(1 - \delta_i)
  \end{flalign}
  \label{eq:mip-form}%
\end{subequations}
with $l_i$ and $u_i$ being lower and upper bounds of $\hat{x_i}$ which can be obtained using interval arithmetic (details in \citep{bunel2018piecewise}).

Our novel contribution is to introduce a method to extend the techniques to handle greedy decoding used in recurrent networks. For a model with greedy decoding, the token with the most likely prediction is fed back as input to the next time step. To implement this mechanism as a mixed integer program, we employ a big-M method \citep{winston2003introduction}:
\begin{subequations}
  \begin{flalign}
  o_{\text{max}} =& \max_{y \in \mathcal{Y}}(o_{y})\span\nonumber\\
  \Rightarrow &\quad o_{\text{max}} \geq o_{y}, \quad \delta_y \in \{0,1\} \quad \forall y \in \mathcal{Y}&\\
  &\quad o_{\text{max}} \leq o_{y} + (\mathbf{u} - l_{y}) (1-\delta_{y}) \quad \forall y \in \mathcal{Y}&\\
  &\quad \sum_{y \in \mathcal{Y}} \delta_{y} = 1&
  \end{flalign}
  \label{eq:maxencoding}
\end{subequations}
with $l_{y}, u_{y}$ being a lower/upper bound on the value of $o_{y}$ and $\mathbf{u}=\max_{y \in \mathcal{Y}} u_{y}$ (these can again be computed using interval arithmetic). Implementing the maximum in this way gives us both a variable representing the value of the maximum ($o_{\text{max}}$), as well as a one-hot encoding of the argmax ($\delta_{y}$). If the embedding for each token is given by \mbox{$\{\text{emb}_i \ | \ i \in \mathcal{Y}\}$}, we can simply encode the input to the following RNN timestep as \mbox{$\sum_{y \in \mathcal{Y}}\delta_y \cdot \text{emb}_y$}, which is a linear function of the variables that we previously constructed.

With this mechanism to encode the greedy decoding, we can now unroll the recurrent model for the desired number of timesteps. To search for an input $x$ with output length $\ell\br{x} \geq \hat{K}$, we unroll the recurrent network for $\hat{K}$ steps and attempt to prove that at each timestep, $\eos$ is not the maximum logit, as in \eqref{eq:greedy-length}. We setup the problem as:
\begin{equation}
% \begin{align}
    \max  \min_{t=1..\hat{K}} \left[\max_{z \neq \textbf{eos}} o_t[z] - o_t[\textbf{eos}]\right]\\
\label{eq:MIP}
% \end{align}
\end{equation}
where $o(k)$ represents the logits in the $k$-th decoding step.  We use an encoding similar to the one of Equation \eqref{eq:maxencoding} to represent the objective function as a linear objective with added constraints. 
If the global optimal value of Eq. \eqref{eq:MIP} is positive, this is a valid counterexample: at all timesteps $t \in [1..\hat{K}]$, there is at least one token greater than the {\bf eos} token, which means that the decoding should continue. On the other hand, if the optimal value is negative, that means that those conditions cannot be satisfied and that it is not possible to generate a sequence of length greater than $\hat{K}$. The {\bf eos} token would necessarily be predicted before. This would imply that our robustness property is True.

\section{Target Model Mechanism}

We use image captioning and machine translation models as specific target examples to study the output length modulation problem. We now introduce their mechanism.

\paragraph{Image captioning models}
% Specifically, for image captioning problem, we use a convolution neural network (CNN) encoder and a recurrent neural network (RNN) decoder \citep{vinyals2015show}. For the machine translation problem, we use a RNN encoder and decoder in the sequence-to-sequence machine framework \citep{sutskever_NIPS2014}.
% \subsection{Image Captioning Models}
The image captioning model we consider is an encoder-decoder model composed of two modules: a convolution neural network (CNN) as an encoder for image feature extraction and a recurrent neural network (RNN) as a decoder for caption generation \citep{vinyals2015show}.

Formally, the input to the model $x$ is an $m\times n$ sized image from the space $\X=[-1,1]^{m\times n}$, the CNN-RNN model computes the output sequence as follows: 
\[
\begin{array}{ll}
     i_0 = \text{CNN}(x); \quad h_0=\mathbf{0}\\
     o_t, h_{t+1} = \text{RNNCell}(i_t, h_t)\\
     y_t=\arg\max(o_t);\quad i_{t+1}=\text{emb}(y_t)
\end{array}
\]
where $\text{emb}$ denotes the embedding function. 

The captioning model first run the input image $x$ through a CNN to obtain the image embedding and feed it to the RNN as the initial input $i_0$ along with the initial state $h_0$. At each decode step, the RNN uses the input $i_t$ and state $h_t$ to compute the new state  $h_{t+1}$ as well as the logits $o_t$ representing the log-probability of the output token distribution in the vocabulary. The output $y_t$ is the token in the vocabulary with highest probability based on $o_t$, and it is embedded into the continuous space using an embedding matrix $W_{emb}$ as $W_{emb}[y_t]$. The embedding is fed to the next RNN cell as the input for the next decoding step.

% In this setting, given an input image $x$, we define the perturbation space $\Sc(x,\delta)$ as the $\ell^\infty$ ball neighbor of the given input image $x$, where the modification of a pixel value in the image $x$ is restricted to $[-\delta, \delta]$. Formally, $\Sc(x,\delta)=\{x'\mid ||x'-x||_\infty \le \delta \land x'\in\X\}$.

% \subsection{Machine Translation Models}
\paragraph{Machine translation models}
The machine translation model is an encoder-decoder model \citep{sutskever2014sequence,cho2014learning} with both the encoder and the decoder being RNNs. Given the vocabulary $V_\text{in}$ of the input language, the valid input space $\X$ is defined as all sentences composed of tokens in $V_\text{in}$, i.e., $\X=\{(x_1,\dots, x_n)\mid x_i\in V, n>0\}$. Given an input sequence $x=(x_1,\dots,x_n)$, the model first calculates its embedding $f(x)$ RNN as follows ($h^e_t$ and $i^e_t$ denote the encoder hidden states and the inputs at the $t$-th time step, respectively. $\text{emb}^e$ denotes the embedding function for each token in the vocabulary). The model then uses $f(x)$ as the initial state $h_0$ for the decoder RNN to generate the output sequence, following the same approach as in the image captioning model.
\vspace{-3pt}
\[
\begin{array}{ll}
    h^e_0 = \mathbf{0};\quad i^e_t = \text{emb}^e(x_t)\\
    h^e_{t} = \text{RNNCell}^e(i^e_t, h^e_{t-1});\quad f(x) = h^e_n
\end{array}
\]
% In this model, we define the $\delta$-perturbation space of a sequence $x=(x_1,\dots,x_n)$ as all sequences of length $n$ with at most $\lceil\delta\cdot n\rceil$ tokens different from $x$ (i.e., $\delta\in[0, 1]$ denotes the percentage of tokens that are allowed to modify). Formally, the perturbation space $\Sc(x,\delta)$ is defined as follows:
% \[\Sc(x,\delta)=\{(x'_1,\dots,x'_n)\mid \sum\limits_{k=1}^{n}\mathbbm{1}[x_k \neq x'_k] < \delta\cdot n\land x'_k\in V\}
% \]
% Since the input vocabulary $V$ is finite, the perturbation space is a finite discrete space with size $|\Sc(x,\delta)|= {n\choose \delta\cdot n}|V|^{\delta\cdot n}$. However, since the vocabulary size is often very large in practice, exhaustively search for the optimum attack is often computationally intractable.

\section{Experiments}
\label{sec:experiment}
We consider the following three models, namely, Multi-MNIST captioning, Show-and-Tell \citep{vinyals2015show}, and Neural Machine Translation (NMT) \citep{sutskever2014sequence,cho2014learning, bahdanau2014neural} models. 

\begin{figure*}
    \centering
    \includegraphics[width=0.8\linewidth]{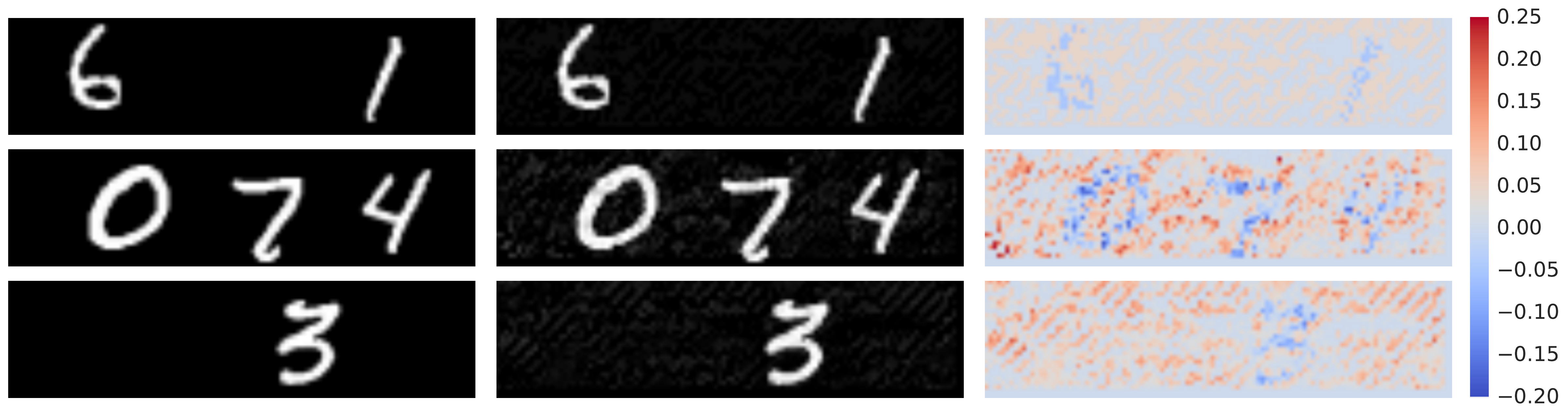}
    \caption{Multi-MNIST examples (left), adversarial examples found by PGD attack (mid), and their differences. For the first group, the model correctly predicts label $l_1=[6, 1]$ on the original image but predicts $l'_1=[6, 1, 1]$ for its corresponding adversarial input. Predictions on the original/adversarial inputs made by model for the second group are $l_2=[0, 7, 4], l'_2=[0, 1, 4, 3]$, and $l_3=[3], l'_3=[3, 3, 5, 3]$ for the third group. The adversarial inputs in the first/second/third groups are found within the perturbation radius $\delta_1=0.1, \delta_2=0.25, \delta_3=0.25$.}
    \label{fig:mmnist_example}
\end{figure*}

\subsection{Details of models and datasets}
\textbf{Multi-MNIST.} The first model we evaluate is a minimal image captioning model for Multi-MNIST dataset. The Multi-MNIST dataset is composed from the MNIST dataset (Figure~\ref{fig:mmnist_example} left). Each image in the dataset is composed from 1-3 MNIST images: each MNIST image (28 * 28) is placed on the canvas of size (28 * 112) with random bias on the $x$-axis. The composition process guarantees that every MNIST image is fully contained in the canvas without overlaps with other images. The label of each image is the list of MNIST digits appearing in the canvas, ordered by their $x$-axis values. The dataset contains 50,000 training images and 10,000 test images, where the training set is constructed from MNIST training set and the test set is constructed from MNIST test set. The images are normalized to $[-1,1]$ before feeding to the captioning model. For this dataset, we train a CNN-RNN model for label prediction. The model encoder is a 4-layers CNN (2 convolution layers and 2 fully connected layers with ReLU activation functions applied in between). The decoder is a RNN with ReLU activation. Both the embedding size and the hidden size are set to 32. We train the model for 300 steps with Adam optimizer based on the cross-entropy loss. The model achieves 91.2\% test accuracy, and all predictions made by the model on the training set have lengths no more than 3.

\textbf{Show-and-Tell.} Show and Tell model \citep{vinyals2015show} is an image captioning model with CNN-RNN encoder-decoder architecture similar to the Multi-MNIST model trained on the MSCOCO 2014 dataset \citep{lin2014microsoft}. Show-and-Tell model uses Inception-v3 as the CNN encoder and an LSTM for caption generation. We use a public version of the pretrained model\footnote{https://github.com/tensorflow/models/} for evaluation. All images are normalized to $[-1, 1]$ and all captions in the dataset are within length 20.

\textbf{NMT.} The machine translation model we study is a Seq2Seq model \citep{sutskever2014sequence, cho2014learning}  with the attention mechanism \citep{bahdanau2014neural} trained on the WMT15 German-English dataset. The model uses byte pair segmentation (BPE) subword units  \citep{DBLP:conf/acl/SennrichHB16a} as vocabulary. The input vocabulary size is $36,548$. The model consists of 4-layer LSTMs of 1024 units with a bidirectional encoder, with the embedding dimension set to 1024. We use a publicly available checkpoint\footnote{https://github.com/tensorflow/nmt} with 27.6 BLEU score on the WMT15 test datasets in our evaluation. At training time, the model restricts the maximum decoding length to 50.

\subsection{Adversarial Attacks}

Our first experiment studies whether adversarial inputs exist for the above models and how they affect model decoding. For each model, we randomly select 100 inputs from the development dataset as attack targets, and compare the output length distributions from random perturbation and PGD attacks.

\paragraph{Multi-MNIST} We evaluate the distribution of output lengths of images with an $\ell^\infty$ perturbation radius of $\delta\in\{0.001, 0.005, 0.01, 0.05, 0.1, 0.5\}$ using both random search and PGD attack. In random search, we generate 10,000 random images within the given perturbation radius for each image in the target dataset as new inputs to the model. In PGD attack, the adversarial inputs are obtained by running 10,000 gradient descent steps with an learning rate of 0.0005 using the Adam Optimizer.

\begin{figure}
    \centering
    \includegraphics[width=\columnwidth]{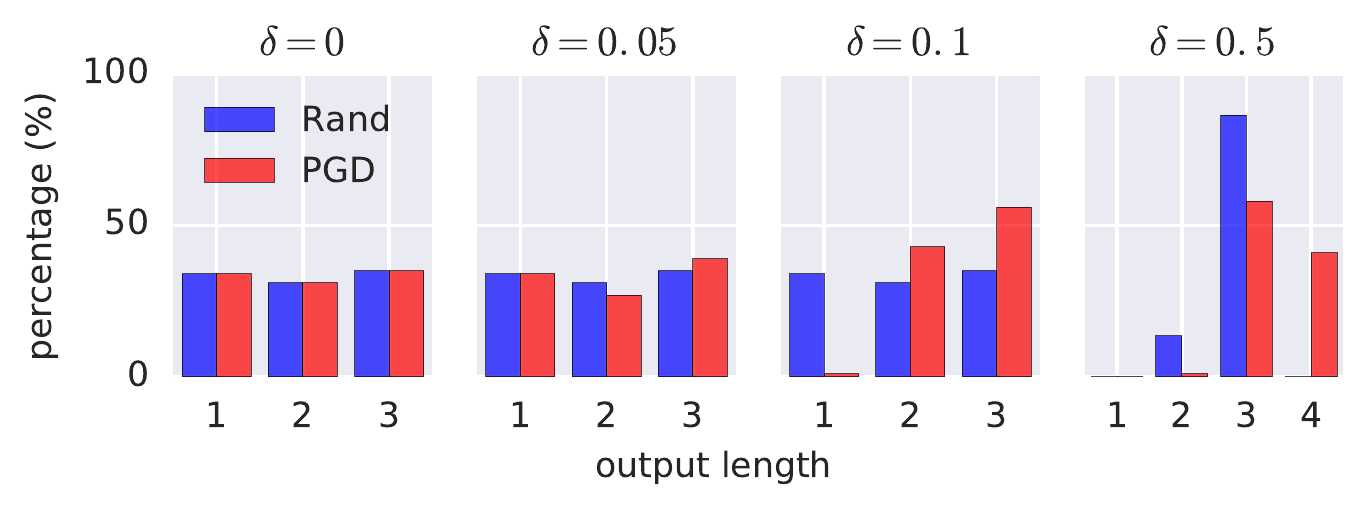}
    \vspace{-15pt}
    \caption{The distribution of output length for random search (denoted as {\bf Rand}) and PGD attack with different perturbation radius $\delta$. The $x$-axis denotes the output length and $y$-axis denotes the number of outputs with the corresponding length. $\delta=0$ (no perturbation allowed) refers to the original output distribution of the target dataset.}
    \label{fig:mmnist_attack}
\end{figure}

Neither of the attack methods can find any adversarial inputs for $\delta\in\{0.001, 0.005, 0.01\}$ perturbation radius (i.e., no perturbation is found for any images in the target dataset within the above $\delta$ to generate an output sequence longer than the original one). Figure~\ref{fig:mmnist_attack} shows the distribution of the output lengths for images with different perturbation radius. Results show that the PGD attack is successful at finding attacks that push the distribution of output lengths higher, particularly at larger values of $\delta$.  Examples of adversarial inputs found by the model are shown in Figure~\ref{fig:mmnist_example}.

\paragraph{Show-and-Tell} For the Show-and-Tell model, we generate attacks within an $\ell^{\infty}$ perturbation radius of $\delta= 0.5$ with both random search and PGD attack on 500 images randomly selected from the development dataset. However, except one adversarial input found by PGD attack that would cause the model to produce an output with size 25, no other adversarial inputs are found that can cause the model to produce outputs longer than 20 words, which is the training length cap. Our analysis shows that the difficulty of attacking the model is resulted from its strong bias on the output sequence distribution and the saturation of sigmoid gates in the LSTM decoder. This result is also consistent with the result found by \cite{show_and_fool} that Show-and-Tell model is ``only able to generate relevant captions learned from the training distribution''.

\paragraph{NMT} We evaluate the NMT model by comparing the output length distribution from adversarial examples generated from random search and PGD attack algorithms. We randomly draw 100 input sentences from the development dataset. The  maximum input length is 78 and their corresponding translations made by the model are all within 75 tokens. We consider the perturbation $\delta\in\{0.3, 0.5, 1.0\}$. 

\vspace{-5pt}
\begin{enumerate}\itemsep-1pt
    \item\emph{Random Search.} In each run of the random attack, given an input sequence with length $n$, we first randomly select $\lceil\delta\cdot n\rceil$ locations to modify, then randomly select substitutions of the tokens at these locations from the input vocabulary, and finally run the NMT model on the modified sequence. We run 10,000 random search steps for the 100 selected inputs, and show the distributions of all outputs obtained from the translation (in the total 1M output sequences).
    \item\emph{PGD Attack.} In PGD attack, we also start by randomly selecting $\lceil\delta\cdot n\rceil$ locations to modify for each input sequence with length $n$. We then run 800 iterations of PGD attack with Adam optimizer using an initial learning rate of 0.005 to find substitutions of the tokens at these selected locations. We plot the output length obtained from running these adversarial inputs through the translation model.
\end{enumerate}
\vspace{-5pt}

\begin{figure}[ht]
    \centering
    \includegraphics[width=\columnwidth]{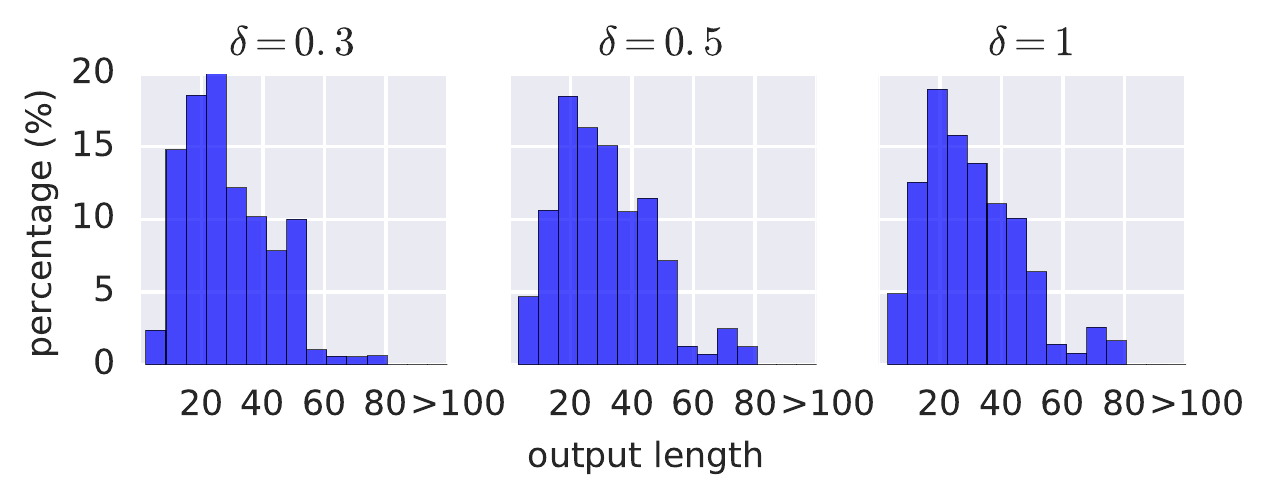}
    \vspace{-15pt}
    \caption{The histogram representing the output length distribution of the NMT model using random search with different perturbations ($\delta \in \{0.3, 0.5, 1\})$. The $x$-axis shows the output length. $y$-axis values are divided by 10,000, the number of random perturbation rounds per image.}
    \label{fig:nmt_random}
\end{figure}

Figure~\ref{fig:nmt_random} shows the distribution of output sequence lengths obtained from random search methods with different $\delta$. We aggregate all sequences with length longer than 100 into the group `$>$100' in the plot. Results show that even random search approach could often craft inputs such that the corresponding output lengths are more than 75 and occasionally generates sentences with output length over 100. The random search algorithm finds 79, 11, 3 for $\delta=$0.3, 0.5, 1, respectively, among the 1M translations that are longer than 100 tokens (at small $\delta$, the search space is more restricted, and random search has a higher success rate of finding long outputs). Notably, the longest sequence found by the random search is a sequence with output length 312 tokens, where the original sequence is only 6.

\begin{figure}[ht]
    \centering
    \includegraphics[width=\columnwidth]{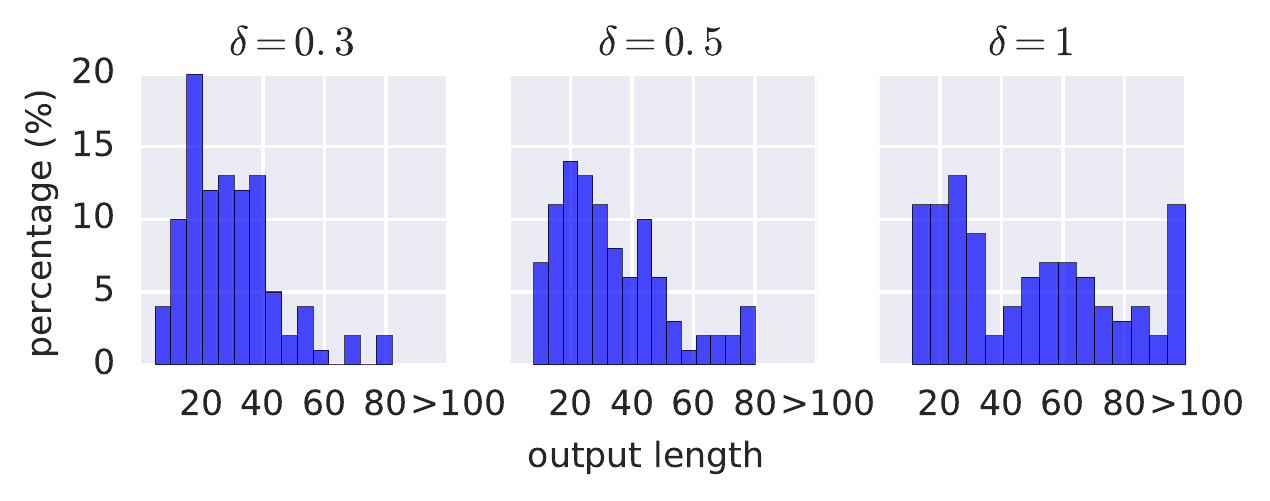}
    \vspace{-15pt}
    \caption{The histogram representing the output length distribution of the NMT model under PGD attack for different $\delta$. $x$-axis shows the output length and $y$-axis shows the number of instances with the corresponding length.}
    \label{fig:nmt_pgd}
\end{figure}

\begin{figure}[ht]
{\footnotesize
{\it
\begin{itemize}[leftmargin=-0.5pt]\itemsep=-1pt
\item[] ($I$) Die Waffe wird ausgestellt und durch den Zaun übergeben.
\item[] ($O$) The weapon is issued and handed over by the fence . \textbf{eos}
\item[] ($I'$) Die namen name descri und ames utt origin i.e. meet grammatisch .
\item[] ($O'$) names name names name names grammatically name names names names names names names names names names names names names names names names names names names names names names names names names names names names names names names names names names names names names names names names names names names names names names names names names names names names names names names names names names names names names names names names names names names names names names names names \textbf{eos}
\end{itemize}}
}
\caption{An example of German to English translation where $I, O$ refer to an original sequence in the dataset and the corresponding translation made by the model. $I', O'$ refer to an adversarial example found by PGD attack and the corresponding model translation.}
\label{fig:nmt_example}
\end{figure}

Figure~\ref{fig:nmt_pgd} shows the result from attacking the NMT model with PGD attack. Results show that PGD attack has relatively low success rate at lower perturbations compared to larger perturbations. With an unconstrained perturbation $\delta=100\%$, PGD attack algorithm discovers more adversarial inputs whose outputs are longer than 100 tokens ($10\%$ among all attacks), which is 1000$\times$ more often than random search. As an extreme case, PGD attack discovered an adversarial input with length 3 whose output length is 575. Examples of adversarial inputs and their corresponding model outputs are shown in Figure~\ref{fig:nmt_example} and the Appendix; we find out that a common feature of the long outputs produced by the translation model is that the output sequences often end with long repetitions of one (or a few) words.   

To analyze the bottleneck of PGD attack on the NMT model, we further run a variation of the PGD attack where the attack space is the (continuous) word embedding space as opposed to the (discrete) token space: we allow the attacker to directly modify token embeddings at selected attack locations to any other vector. PGD attack on this variation achieves a 100\% success rate to find adversarial token embeddings such that the model outputs are longer than 500 tokens. This indicates that the discrete space is a bottleneck for consistently finding stronger attacks.

\subsection{Verification}
%\djred{Show bounds on output length based on verification vs worst output-lengthening attacks found by attack algorithm, as a function of perturbation radius on input image.}

Our implementation of the verification algorithm using the mixed integer programming \eqref{eq:MIP} is implemented using SCIP \citep{GleixnerScip}. We run our verification algorithm on the Multi-MNIST dataset, attempting to formally prove the robustness of the model to attacks attempting to generate an output longer than the ground truth. For each input image, we set a timeout of 30 minutes for the solver. 

\begin{figure}
    \centering
    \includegraphics[width=0.65\columnwidth]{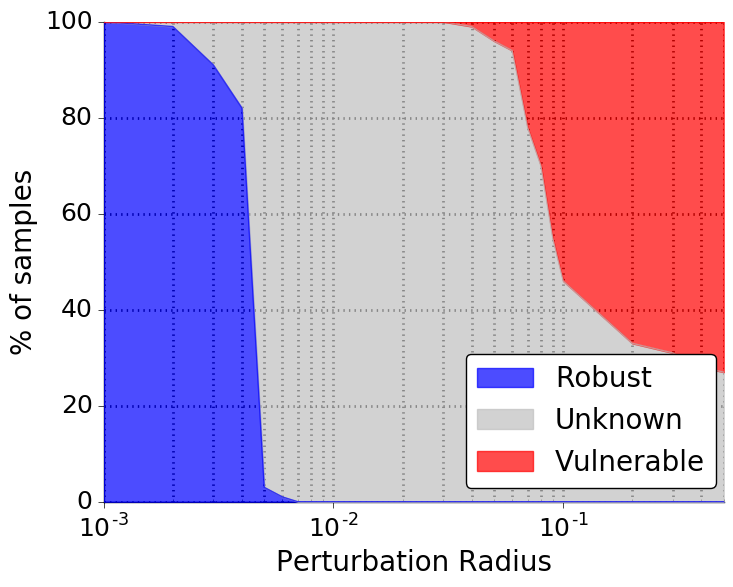}
    \caption{Proportion of samples being provably robust (in blue), vulnerable (in red) or of unknown robustness status (in gray) to an attack attempting to make the model generate an output sequence longer than the ground truth, as a function of the perturbation radius allowed to the attacker. For small radiuses, the MIP can prove that no attacks can be successful. For large radiuses, we are able to find successful attacks. }
    \label{fig:verif_stacked}
\end{figure}

The results in Figure \ref{fig:verif_stacked} show that our verification algorithm is able to verify formally that no attacks exists for small perturbation radiuses. However, as the perturbation radius increases, our verification algorithm times out and is not able to explore the full space of valid perturbations and thus cannot decide whether attacks exists in the given space. For this reason, the number of robust samples we report is only a lower bound on the actual number of robust samples. Conversely, the vulnerable samples that we exhibit give us an upper bound on the number of those robust samples. As shown by the large proportion of samples of unknown status, there is currently still a gap between the capabilities of formal verification method and attacks.

\section{Conclusion}
In this paper, we introduce the existence and the construction of the output-length modulation problem. 
We propose a differentiable proxy that can be used with PGD to efficiently find output-lengthening inputs. 
We also develop a verification approach to formally prove certain models cannot produce outputs greater than a certain length. 
We show that the proposed algorithm can produce adversarial examples that are 50 times longer than the input for machine translation models, and the image-captioning model can conform the output size is less than certain maximum length using the verification approach.
In future work, we plan to study adversarial training of sequential output models using the generated attacks, to models that are robust against output lengthening attacks, and further, verify this formally.

% (a) find adversarial examplesi.e.search for inputs that lead the model to produce out-puts with a larger length and thus show that the modelunder consideration violates the specification,
% and 
% (b) perform formal verification {\em i.e.} prove that the model under consideration will be consistent with the specification for the given range of inputs, if such attacks are indeed impossible.

{\small
\bibliographystyle{ieee_fullname}
\bibliography{references}
}

\clearpage

\appendix

\section{NMT Examples}

We show a few more examples of the NMT model. We use $I_k,O_k$ to refer to original inputs and their translations produced by the model, and $I'_k, O'_k$ for adversarial inputs / outputs. The following adversarial examples are found by with $\delta=1$ (unconstrained perturbation in the discrete token space).

{\it
\begin{itemize}[leftmargin=-0.5pt]\itemsep=-1pt
\item[] ($I_1$) Eink-aufen vom So-fa aus
\item[] ($O_1$) Shopping from the sofa \textbf{eos}
\item[] ($I_1'$) Schlusselwortern Anreise nähe Hotelsafe kes hotel
\item[] ($O_1'$) 
keyword 4 Com-city Hotel near Safety Deposit Box kes hotel kes garage kes kes kes kes kes kes kes kes kes kes kes kes kes hotel kes kes next street kes kes garage kes kes hotel kes kes kes kes kes kes kes Hotel kes kes kes kes kes kes Hotel kes kes kes kes kes kes kes kes kes kes kes kes kes kes kes kes kes kes kes kes kes kes kes kes kes \textbf{eos}
\end{itemize}}

% {\it
% \begin{itemize}[leftmargin=-0.5pt]\itemsep=-1pt
% \item[] ($I_2$) Die Preise und das War-ens-or-timent machten Wal-mar-t zu einem der größten Unternehmen in den USA .
% \item[] ($O_2$) The prices and assor-tment made Wal-mar-t one of the largest companies in the United States . \textbf{eos}

% \item[] ($I_2'$) downlo-muse-tribute draft quantities Slideshows main nehmbar clean eingebettet injured behindertenfreundliche kommend tücher leug-diesbezü-fur overview ← wake Pakist-

% \item[] ($O_2'$) 
% Download the text of the project 's main features include tips for access to event-sites related to Pakistan Pakistan Pakistan Pakistan Pakistan Pakistan Pakistan Pakistan Pakistan Pakistan Pakistan Pakistan Pakistan Pakistan Pakistan Pakistan Pakistan Pakistan Pakistan Pakistan Pakistan Pakistan Pakistan Pakistan Pakistan Pakistan Pakistan Pakistan Pakistan Pakistan Pakistan Pakistan Pakistan Pakistan Pakistan Pakistan Pakistan Pakistan Pakistan Pakistan Pakistan Pakistan Pakistan Pakistan Pakistan Pakistan Pakistan Pakistan Pakistan Pakistan Pakistan Pakistan Pakistan Pakistan Pakistan Pakistan Pakistan Pakistan Pakistan Pakistan Pakistan Pakistan Pakistan Pakistan Pakistan Pakistan Pakistan Pakistan Pakistan Pakistan \textbf{eos}
% \end{itemize}}

{\it
\begin{itemize}[leftmargin=-0.5pt]\itemsep=-1pt
\item[] ($I_2$) Un-zufrie-dene
\item[] ($O_2$) Related dis-satisfied \textbf{eos}
\item[] ($I_2'$) Anmeldung aspects ICEcat
\item[] ($O_2'$) Regi- ment aspects ICEcat ICEcat ICEcat ICEcat ICEcat ICEcat ICEcat ICEcat ICEcat ICEcat ICEcat ICEcat ICEcat ICEcat ICEcat ICEcat ICEcat ICEcat ICEcat ICEcat ICEcat ICEcat ICEcat ICEcat ICEcat ICEcat ICEcat ICEcat ICEcat ICEcat ICEcat ICEcat ICEcat ICEcat ICEcat ICEcat ICEcat ICEcat ICEcat ICEcat ICEcat ICEcat ICEcat ICEcat ICEcat ICEcat ICEcat ICEcat ICEcat ICEcat ICEcat ICEcat ICEcat ICEcat ICEcat ICEcat ICEcat ICEcat ICEcat ICEcat ICEcat ICEcat ICEcat ICEcat ICEcat ICEcat ICEcat ICEcat ICEcat ICEcat ICEcat ICEcat ICEcat ICEcat ICEcat ICEcat ICEcat ICEcat ICEcat ICEcat ICEcat ICEcat ICEcat ICEcat ICEcat ICEcat ICEcat ICEcat ICEcat ICEcat ICEcat ICEcat ICEcat ICEcat ICEcat ICEcat ICEcat ICEcat ICEcat ICEcat ICEcat ICEcat ICEcat ICEcat ICEcat ICEcat ICEcat ICEcat ICEcat ICEcat ICEcat ICEcat ICEcat ICEcat ICEcat ICEcat ICEcat ICEcat ICEcat ICEcat ICEcat ICEcat ICEcat ICEcat ICEcat ICEcat ICEcat ICEcat ICEcat ICEcat ICEcat ICEcat ICEcat ICEcat ICEcat ICEcat ICEcat ICEcat ICEcat ICEcat ICEcat ICEcat ICEcat ICEcat ICEcat ICEcat ICEcat ICEcat ICEcat ICEcat ICEcat ICEcat ICEcat ICEcat ICEcat ICEcat ICEcat ICEcat ICEcat ICEcat ICEcat ICEcat ICEcat ICEcat ICEcat ICEcat ICEcat ICEcat ICEcat ICEcat ICEcat ICEcat ICEcat ICEcat ICEcat ICEcat ICEcat ICEcat ICEcat ICEcat ICEcat ICEcat ICEcat ICEcat ICEcat ICEcat ICEcat ICEcat ICEcat ICEcat ICEcat ICEcat ICEcat ICEcat ICEcat ICEcat ICEcat ICEcat ICEcat ICEcat ICEcat ICEcat ICEcat ICEcat ICEcat ICEcat ICEcat ICEcat ICEcat ICEcat ICEcat ICEcat ICEcat ICEcat ICEcat ICEcat ICEcat ICEcat ICEcat ICEcat ICEcat ICEcat ICEcat ICEcat ICEcat ICEcat ICEcat ICEcat ICEcat ICEcat ICEcat ICEcat ICEcat ICEcat ICEcat ICEcat ICEcat ICEcat ICEcat ICEcat ICEcat ICEcat ICEcat ICEcat ICEcat ICEcat ICEcat ICEcat ICEcat ICEcat ICEcat ICEcat ICEcat ICEcat ICEcat ICEcat ICEcat ICEcat ICEcat ICEcat ICEcat ICEcat ICEcat ICEcat ICEcat ICEcat ICEcat ICEcat ICEcat ICEcat ICEcat ICEcat ICEcat ICEcat ICEcat ICEcat ICEcat ICEcat ICEcat ICEcat ICEcat ICEcat ICEcat ICEcat ICEcat ICEcat ICEcat ICEcat ICEcat ICEcat ICEcat ICEcat ICEcat ICEcat ICEcat ICEcat ICEcat ICEcat ICEcat ICEcat ICEcat ICEcat ICEcat ICEcat ICEcat ICEcat ICEcat ICEcat ICEcat ICEcat ICEcat ICEcat ICEcat ICEcat ICEcat ICEcat ICEcat ICEcat ICEcat ICEcat ICEcat ICEcat ICEcat ICEcat ICEcat ICEcat ICEcat ICEcat ICEcat ICEcat ICEcat ICEcat ICEcat ICEcat ICEcat ICEcat ICEcat ICEcat ICEcat ICEcat ICEcat ICEcat ICEcat ICEcat ICEcat ICEcat ICEcat ICEcat ICEcat ICEcat ICEcat ICEcat ICEcat ICEcat ICEcat ICEcat ICEcat ICEcat ICEcat ICEcat ICEcat ICEcat ICEcat ICEcat ICEcat ICEcat ICEcat ICEcat ICEcat ICEcat ICEcat ICEcat ICEcat ICEcat ICEcat ICEcat ICEcat ICEcat ICEcat ICEcat ICEcat ICEcat ICEcat ICEcat ICEcat ICEcat ICEcat ICEcat ICEcat ICEcat ICEcat ICEcat ICEcat ICEcat ICEcat ICEcat ICEcat ICEcat ICEcat ICEcat ICEcat ICEcat ICEcat ICEcat ICEcat ICEcat ICEcat ICEcat ICEcat ICEcat ICEcat ICEcat ICEcat ICEcat ICEcat ICEcat ICEcat ICEcat ICEcat ICEcat ICEcat ICEcat ICEcat ICEcat ICEcat ICEcat ICEcat ICEcat ICEcat ICEcat ICEcat ICEcat ICEcat ICEcat ICEcat ICEcat ICEcat ICEcat ICEcat ICEcat ICEcat ICEcat ICEcat ICEcat ICEcat ICEcat ICEcat ICEcat ICEcat ICEcat ICEcat ICEcat ICEcat ICEcat ICEcat ICEcat ICEcat ICEcat ICEcat ICEcat ICEcat ICEcat ICEcat ICEcat ICEcat ICEcat ICEcat ICEcat ICEcat ICEcat ICEcat ICEcat ICEcat ICEcat ICEcat ICEcat ICEcat ICEcat ICEcat ICEcat ICEcat ICEcat ICEcat ICEcat ICEcat ICEcat ICEcat ICEcat ICEcat ICEcat ICEcat ICEcat ICEcat ICEcat ICEcat ICEcat ICEcat ICEcat ICEcat ICEcat ICEcat ICEcat ICEcat ICEcat ICEcat ICEcat ICEcat ICEcat ICEcat ICEcat ICEcat ICEcat ICEcat ICEcat ICEcat ICEcat ICEcat ICEcat ICEcat ICEcat ICEcat ICEcat ICEcat ICEcat ICEcat ICEcat ICEcat ICEcat ICEcat ICEcat ICEcat ICEcat ICEcat ICEcat ICEcat ICEcat ICEcat ICEcat ICEcat ICEcat ICEcat ICEcat ICEcat ICEcat ICEcat ICEcat ICEcat ICEcat ICEcat ICEcat ICEcat ICEcat ICEcat ICEcat ICEcat ICEcat ICEcat ICEcat ICEcat ICEcat ICEcat ICEcat ICEcat ICEcat ICEcat ICEcat ICEcat ICEcat ICEcat ICEcat ICEcat \textbf{eos}
\end{itemize}}

\clearpage

\begin{figure*}
  \centering
  \begin{tikzpicture}

  \newlength{\probwidth}
  \setlength{\probwidth}{1em}
  \newlength{\probheight}
  \setlength{\probheight}{1em}

  \tikzset{
    dummy/.style={inner sep=0pt, outer sep=0pt},
    empty_cell/.style={draw=black, inner sep=0pt, outer sep=0pt},
    cell_container/.style={empty_cell, minimum width=\probwidth, minimum height=\probheight},
    cell_content/.style={empty_cell, minimum width=\probwidth,anchor=south},
    arr/.style={-{Latex[width=1em,length=0.5em]}},
    instr_box/.style={draw, minimum width=4\probwidth},
    opbox/.style={draw, inner sep=0.3em},
  }

  \newcommand{\probaCell}[1]{
    % #1 - Arguments for the container
    \path node[cell_container,#1]{};
 }

 \newcommand{\probaRow}[4]{
   % #1 - Numbers of cells
   % #2 - Prefix name
   % #3 - Common content for the proba cells
   % #4 - Specific content for the first proba cell
   \probaCell{name=#2-1, #3, #4}
   \foreach \colidx in {2,...,#1}{
     \pgfmathsetmacro\prevcol{int(\colidx - 1)}
     %\node[draw] at (\colidx , \prevcol ){\colidx, \prevcol};
     \probaCell{name=#2-\colidx, right = 0 of #2-\prevcol, #3}
   }
   \node[fit=(#2-1) (#2-#1), name={#2},inner sep=0pt, outer sep=0pt]{};
 }

 \newcommand{\probaCol}[4]{
   % #1 - Numbers of cells
   % #2 - Prefix name
   % #3 - Common content for the proba cells
   % #4 - Specific content for the first proba cell
   \probaCell{name=#2-1, #3, #4}
   \foreach \colidx in {2,...,#1}{
     \pgfmathsetmacro\prevcol{int(\colidx - 1)}
     %\node[draw] at (\colidx , \prevcol ){\colidx, \prevcol};
     \probaCell{name=#2-\colidx, below = 0 of #2-\prevcol, #3}
   }
   \node[fit=(#2-1) (#2-#1), name={#2},inner sep=0pt, outer sep=0pt]{};
 }

 \tikzset{neuron/.style={draw, circle,inner sep=0, outer sep=0, minimum size=0.5cm}};
 \tikzset{dummy/.style= {inner sep=0, outer sep=0}}

 % Inputs
 \probaRow{2}{inputImg}{fill=black}{fill=white}
 \node[dummy, text=black, below = 0 of inputImg]{\large $x$};

 % Build the CNN
 \node[neuron, above left = 0.7 of inputImg-1](reluA){};
 \draw[black, thick](reluA.west) to (reluA.center) to (reluA.north east);
 \node[above left = 0.01 of reluA.north west](a){\large $a$};
 \node[below left = 0.01 of reluA.south west](ahat){\large $\hat{a}$};
 \node[neuron, above right = 0.7 of inputImg-2](reluB){};
 \draw[black, thick](reluB.west) to (reluB.center) to (reluB.north east);
 \node[above right = 0.01 of reluB.north east](b){\large $b$};
 \node[below right = 0.01 of reluB.south east](bhat){\large $\hat{b}$};

 % CNN output
 \probaRow{3}{cnnEmb}{fill=orange}{above = 1.5 of inputImg.north west};
 \node[dummy, text=orange, left = 0 of cnnEmb]{\large $i_0$};

 % Draw the connection of the CNN - Input layer
 \foreach \inpidx in {1,...,2}{
   \draw[-latex] (inputImg-\inpidx.north) to (reluA.south);
   \draw[-latex] (inputImg-\inpidx.north) to (reluB.south);
 }
 % Draw the connection of the CNN - Output layer
 \foreach \outidx in {1,...,3}{
   \draw[-latex] (reluA.north) to (cnnEmb-\outidx.south);
   \draw[-latex] (reluB.north) to (cnnEmb-\outidx.south);
 }

 % Initial value for h_0
 \probaRow{3}{h0}{fill=blue}{above = 1 of cnnEmb-1.north}
 \node[dummy, text=blue, above = 0 of h0]{\large $h_0 = 0$};

 % Duplicate for inputs to the RNN
 \probaRow{3}{i0}{fill=orange}{right = 2 of cnnEmb}
 \probaRow{3}{h0cop}{fill=blue}{right = 0 of i0.east}
 % Arrows to link it
 \draw[->, dashed, color=orange] (cnnEmb.east) to (i0.west);
 \draw[->, dashed, color=blue] (h0.south east) to (h0cop.north west);

 % Output of the first RNN Cell
 \node[draw, thick, above = of i0.east](rnnCell0) {RNNCell};
 \probaRow{2}{o0}{fill=green}{above = of rnnCell0};
 \node[dummy, text=green, left = 0 of o0]{\large $o_0$};
 \probaRow{3}{h1}{fill=blue, opacity=0.7}{above right = 0.5 of rnnCell0};
 \node[dummy, text=blue, above = 0 of h1]{\large $h_1$};

 % Connection for RNNCell
 \draw[-](i0.north west) to (rnnCell0.south west);
 \draw[-](h0cop.north east) to (rnnCell0.south east);
 \draw[-](rnnCell0.north west) to (o0.south west);
 \draw[-](rnnCell0.north east) to (o0.south east);
 \draw[-](rnnCell0.north west) to (h1.south west);
 \draw[-](rnnCell0.north east) to (h1.south east);

 % Inputs to the second RNN Cell
 \node[dummy](anchor2) at (inputImg.north -| h1){};
 \probaCol{2}{delta1}{fill=green, dashed}{right = 2 of anchor2};
 \node[dummy, text=green, left = 0 of delta1]{\large $\delta_1$};
 \draw[->, dashed, color=green](o0) to (delta1);
 \node[right = 0 of delta1](mult){\Large $\times$};
 \probaRow{3}{rnnEmb-0}{fill=red}{right = 0.6 of delta1-1}
 \probaRow{3}{rnnEmb-1}{fill=red}{right = 0.6 of delta1-2}
 \node[dummy, text=red, right = 0 of rnnEmb-0]{\large emb};

 \node[dummy](rnn2inp) at (h0cop.south -| mult){};
 \probaRow{3}{i1}{fill=red, opacity=0.8}{above = 0 of rnn2inp}
 \node[dummy, text=red, left = 0 of i1]{\large $i_1$};
 \draw[->, dashed, color=red](mult) to (i1.south);
 \probaRow{3}{h1cop}{fill=blue, opacity=0.7}{right = 0 of i1.east}
 \draw[->, dashed, color=blue](h1) to (h1cop.north);

 % Second RNN Cell
 \node[draw, thick, above = of i1.east](rnnCell1) {RNNCell};
 \draw[-](i1.north west) to (rnnCell1.south west);
 \draw[-](h1cop.north east) to (rnnCell1.south east);
 \probaRow{2}{o1}{fill=green}{above = of rnnCell1};
 \node[dummy, text=green, left = 0 of o1]{\large $o_1$};
 \probaRow{3}{h2}{fill=blue, opacity=0.7}{above right = 0.5 of rnnCell1};
 \node[dummy, text=blue, above = 0 of h2]{\large $h_2$};
 \draw[-](rnnCell1.north west) to (o1.south west);
 \draw[-](rnnCell1.north east) to (o1.south east);
 \draw[-](rnnCell1.north west) to (h2.south west);
 \draw[-](rnnCell1.north east) to (h2.south east);

 \node[right = 3 of rnnCell1]{\LARGE $\dots$};
 \node[dummy](anchor3) at (inputImg.north -| h2){};
 \probaCol{2}{delta2}{fill=green, dashed}{right = of anchor3};
 \draw[->, dashed, color=green](o1) to (delta2);
 \node[dummy, text=green, left = 0 of delta2]{\large $\delta_2$};

 %% Draw bounding box for explanation groups
 % Draw the bounding box around the embedding part
 \node[draw, inner sep=\probheight, very thick,
 fit=(inputImg) (reluA) (reluB) (cnnEmb)](cnnBox){};
 \node[below = 0 of cnnBox.south]{\Large (a) CNN embedding};

 \node[draw, inner sep=\probheight, very thick,
 fit=(i0) (h0cop) (o0) (h1)](rnnCellBox){};
 \node[above = 0 of rnnCellBox.north]{\Large  (b) RNN Cell};

 \node[draw, inner sep=\probheight, very thick,
 fit=(i1) (rnnEmb-0) (rnnEmb-1) (delta1)](embBox){};
 \node[below = 0 of embBox.south]{\Large (d) Vocab embedding};

 \path let
         \p1 = ($(o1.north west)-(delta2.south east)$),
         \n1 = {veclen(\p1)}
         in
         (o1.north west) -- (delta2.south east)
         node[midway, sloped, draw, ellipse,
              minimum width=\n1, minimum height=0.4*\n1] {};
 \node[above = 0 of o1.north]{\Large (c) Argmax};

\end{tikzpicture}
	\caption{Illustration of a toy model like the one we used for the
    Multi-MNIST application.
    (a) The input image is composed of two pixels, the
    image embedding network has one hidden layer with two neurons $a$ and $b$,
    from which the inital input $i_0$ to the network is generated.
    (b) At each time step, the input to the network $i_k$ is combined (here simply
    concatenated) with the state $h_k$, based on which this step's output $o_k$
    and the next stat $h_{k+1}$ are computed.
    (c) The maximum output of $o_k$ is identified and fed back into the network.
    (d) The input of the next step of the network is the vocabulary embedding
    corresponding to the maximum output.
  }
	\label{fig:rnn_example_fig}
\end{figure*}
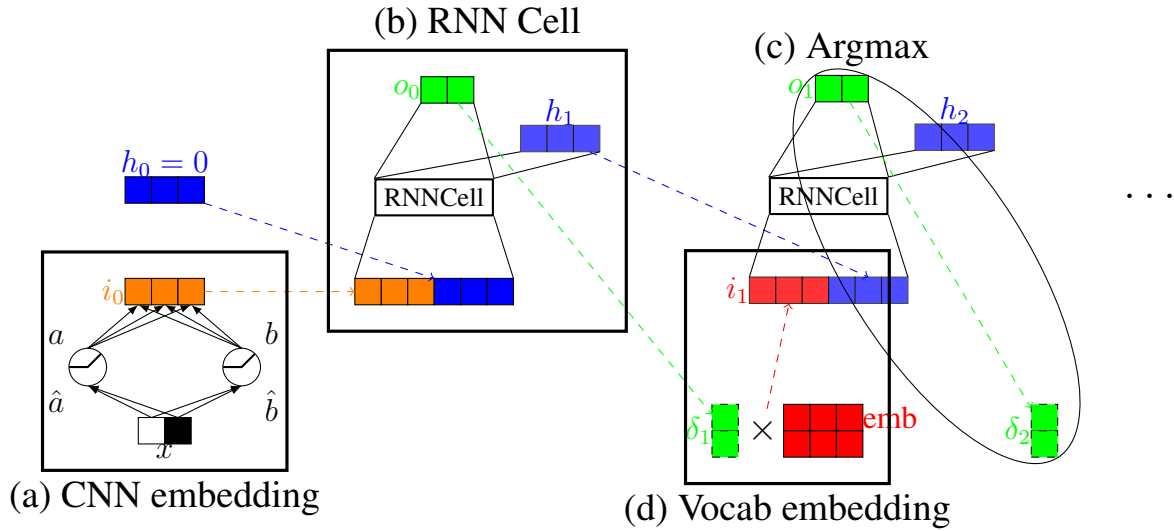

\section{A Verification Toy Example}
We now provide an example of how a formal verification problem would be encoded
according to the methodology described in Section 4. To do so, we take the
illustrative example of Figure~\ref{fig:rnn_example_fig} and run through the
different types of variables and constraints that needs to be defined to encode
the problem. For clarity, we will represent all solver variables with an underline.

\newcommand{\und}[1]{\underline{#1}}

\subsection{CNN Embedding Encoding}
The input domain $\mathcal{S}(x, \delta)$ would be initially defined by a set of bound constraints over the inputs. In the case of $\mathcal{L}_\infty$ adversarial perturbation
around an image $x_{\text{ref}}$, this would take the form of defining
\mbox{$l_0 = x_{\text{ref}} - \delta$} and \mbox{$u_0 = x_{\text{ref}} +
  \delta$}, and creating a variable $x$, that will be subjected to the
constraints:
\begin{equation}
  \begin{split}
    \und{x} \geq l_0\\
    \und{x} \leq u_0
  \end{split}
\end{equation}
We would then need to create some variables corresponding to the input of the
hidden layer, which would be constrained by a linear equality depending on the
network input variable, and the network weights $W_1$
\begin{equation}
  \begin{split}
    \und{\hat{a}} = W_{1,a} \und{x} + b_{1,a} \qquad
    \und{\hat{b}} = W_{1,b} \und{x} + b_{1,b}
  \end{split}
  \label{eq:example-linear}
\end{equation}
The next step consists in creating variables representing the output of the
hidden layer, after the ReLU activation function. In order to do so, we need to
first obtain bounds on $\und{\hat{a}}$ and $\und{\hat{b}}$. Interval analysis is
one of the way to achieve this:
\begin{equation*}
  \begin{split}
    l_a = W_{1,a}^+ l_0 + W_{1,a}^- u_0 + b_{1, a} \\
    u_a = W_{1,a}^+ u_0 + W_{1,a}^- l_0 + b_{1, a},
  \end{split}
\end{equation*}
which bounds on $\und{\hat{b}}$ being obtained similarly.

Using those bounds, we can now generate variables representing the ReLU outputs,
and constrain them to only take the appropriate values. Using the formulation
described in Equation~\eqref{eq:mip-form}, the constraints required for the ReLU
A are:
\begin{equation}
  \begin{split}
    &\und{\delta_a} \in \{0, 1\}\\
    &\und{a} \geq 0\\
    &\und{a} \geq \und{\hat{a}}\\
    &\und{a} \leq u_a \und{\delta_a}\\
    &\und{a} \leq \und{\hat{a}} - l_a \left( 1 - \und{\delta_a} \right)
  \end{split}
  \label{eq:example-relu}
\end{equation}
The outputs of the embedding network can be constructed in a manner similar to
the way we constructed the input to the hidden layer, by enforcing a single linear
equality constraint:
\begin{equation}
  \und{i_0} = W_2 \left[ \begin{array}{l}\und{a}\\\und{b}\end{array} \right] + b_2
  \label{eq:example-linear2}
\end{equation}
We can then easily derive bounds $\left(l_{i_0}, u_{i_0}\right)$ by using the
same interval analysis method from before.

We have now seen how to represent the allowed images, as well as the
corresponding activations throughout the embedding network, as a set of
variables and the linear constraints that they should satisfy. We will now move
on to explaining how the decoder recurrent network is encoded into MIP constraints.

\subsection{RNN Cell Encoding}
The encoding of the recurrent cell is similar to the CNN embedding
described in the previous section.

Based on an initial inputs $\left( \und{i_k}, \und{h_k}\right)$, for which we
have bounds and constraints, we will create new variables representing the
RNN cell activations corresponding to those inputs. Our example doesn't
specifically show how the RNN cell is implemented, but provided that it is
composed of only linear operations (which will give constraints in the style of
Equations~\eqref{eq:example-linear} and \eqref{eq:example-linear2}) or ReLU
operations (with constraints in the style of Equations~\eqref{eq:example-relu}),
the previously described techniques will be able to generate constrained
variables $\left( \und{o_k}, \und{h_{k+1}}\right)$, as well as upper and lower
bound for those variables.

\subsection{Argmax Encoding}
We will now show how we can create a variable that represents the argmax of the
RNN cell output. In our example, the size of the vocabulary is 2, but the
technique, described in Equation~\ref{eq:maxencoding}, is valid beyond that.

Assuming we have already created the variables $\und{o_k}$, and we have access
to their bounds $\left( l_{o_k}, u_{o_k} \right)$, we will create the new
variables $\und{o_{k, \text{max}}}$ and $\und{\delta_{k}}$, subject to the constraints:
\begin{equation}
  \begin{split}
    & \und{\delta_k[0]} \in \{0, 1\} \qquad \qquad \und{\delta_k[1]} \in \{0, 1\} \\
    &\und{o_{k, \text{max}}} \geq \und{o_k[0]} \qquad \qquad  \und{o_{k, \text{max}}} \geq \und{o_k[1]}\\
    &\und{o_{k, \text{max}}} \leq \und{o_k[0]} + \left(\max\left[u_{o_k}\right] - l_{o_k}[0]\right) \left( 1 - \und{\delta_k[0]} \right)\\
    &\und{o_{k, \text{max}}} \leq \und{o_k[1]} + \left(\max\left[u_{o_k}\right] - l_{o_k}[1]\right) \left( 1 - \und{\delta_k[1]} \right)\\
    &\und{\delta_k[0]} + \und{\delta_k[1]} = 1
  \end{split}
\end{equation}
The value $\und{o_{k, \text{max}}}$ can be used when we need to encode the
values of the maximum (such as in the inner term of Equation~\eqref{eq:MIP}),
and we will be able to use the argmax variable $\und{\delta_k}$ in order to
implement the greedy decoding mechanism.

\subsection{Vocabulary Embedding}
We need to generate the variable $\und{i_{k+1}}$ that corresponds to the
embedding of the token that was last predicted by the network. We have access to
the $\und{\delta_k}$ variable, which is by construction a one-hot binary
variable indicating which was the generated token. We can simply encode:
\begin{equation}
  \und{i_{k+1}} = \sum_{y=1..2} \und{\delta_{k}[y]} \times \text{emb}[y]
  \label{eq:example-vocembedding-encoding}
\end{equation}
Given that only one of the $\und{\delta_{k}[y]}$ is going to be equal to 1, all
the other being 0, this will exactly correspond to choosing the embedding
corresponding to the selected token.

Note that in order to continue building the Mixed Integer Programming
problem, we need to have access to lower and upper bounds on the values taken by
$\und{i_{k+1}}$. These can be simply obtained by taking a maximum / minimum
along the token dimension of the token embeddings:
\begin{equation*}
  \begin{split}
    l_{i_{k+1}} = \min_{y=1..2} \text{emb}[y]\\
    u_{i_{k+1}} = \max_{y=1..2} \text{emb}[y]
  \end{split}
\end{equation*}

Using these bounds, as well as the derived bounds on $\und{h_{k+1}}$ that came
from propagating bounds through the RNN cell, we now have variables and bounds
for the next step of RNN enrolling. This enrolling can be repeated until the
target length to obtain a MIP encoding of the sequence generated by the network.

Inputting all these constraints into a Mixed Integer Programming solver will
then generate the desired formal verification result.

\end{document}